\documentclass[sigconf]{acmart}

\usepackage{subfigure}
\usepackage{multirow}

\AtBeginDocument{%
  \providecommand\BibTeX{{%
    \normalfont B\kern-0.5em{\scshape i\kern-0.25em b}\kern-0.8em\TeX}}}

\setcopyright{acmcopyright}
\acmDOI{XXXXXXX.XXXXXXX}

\acmConference[Preprint]{}{}{}

\acmISBN{978-1-4503-XXXX-X/18/06}
\newcommand{\method}{\textsc{MELODY}}
\newcommand{\fe}{\textsc{OFE}}
\newcommand{\ad}{\textsc{SemiAD}}
\newcommand{\mat}[1]{{\textbf #1}}

\begin{document}

\title[MELODY: Robust Semi-Supervised Hybrid Model for Entity-Level Online Anomaly Detection]{MELODY: Robust Semi-Supervised Hybrid Model for Entity-Level Online Anomaly Detection with Multivariate Time Series}

\author{Jingchao Ni, Gauthier Guinet, Peihong Jiang$^{*}$, Laurent Callot, Andrey Kan}\authornote{Independent researcher, work conducted while employed at Amazon}
\affiliation{AWS AI Labs\\
\{jingchni, guinetgg, lcallot, avkan\}@amazon.com, peihong.jiang.7@gmail.com
\country{}
}

\renewcommand{\shortauthors}{Ni et al.}

\begin{abstract}

% \lcallot{}{I've reworked the abstract. Please check if you are satistifed.}

% \avkan{}{How about calling the method MELODY (Model for Entity-Level Online anomaly Detection? If you prefer the current name, consider SHMAD instead of SHM-AD.}

% \avkan{}{Also see if you can organically add words such as ``production'', ``large scale'', etc. to the title}

% \guinetgg{}{Following Andrey's comment, perhaps large scale is better suited than robust ? The latter doesn't directly map to a given part of the paper, or said }

In large IT systems, software deployment is a crucial process in online services as their code is regularly updated. However, a faulty code change may degrade the target service's performance and cause cascading outages in downstream services. Thus, software deployments should be comprehensively monitored, and their anomalies should be detected timely. %Although many anomaly detection techniques have been proposed for multivariate time series (MTS), few of them can be readily used for online services in industrial environments, which require new methods for entity-level ({\em e.g.}, service) anomaly detection.
In this paper, we study the problem of anomaly detection for deployments. We begin by identifying the challenges unique to this anomaly detection problem, which is at entity-level ({\em e.g.}, deployments), relative to the more typical problem of anomaly detection in multivariate time series (MTS). The unique challenges include the heterogeneity of deployments, the low latency tolerance, the ambiguous anomaly definition, and the limited supervision. To address them, we propose a novel framework, semi-supervised hybrid \underline{M}odel for \underline{E}ntity-\underline{L}evel \underline{O}nline \underline{D}etection of anomal\underline{Y} (\method). \method\ first transforms the MTS of different entities to the same feature space by an online feature extractor, then uses a newly proposed semi-supervised deep one-class model for detecting anomalous entities. We evaluated \method\ on real data of cloud services with 1.2M+ time series. %collected in 9 months.
The relative F1 score improvement of \method\ over the state-of-the-art methods ranges from 7.6\% to 56.5\%. The user evaluation suggests \method\ is suitable for monitoring deployments in large online systems.

\end{abstract}


%\ccsdesc[100]{Do Not Use This Code~Generate the Correct Terms for Your Paper}

%\keywords{Do, Not, Us, This, Code, Put, the, Correct, Terms, for,
%  Your, Paper}



\maketitle

\section{Introduction}\label{sec.intro}

As cloud native systems become prevalent in modern IT industry, most applications ``born on cloud'' are composed of a multitude of interconnected services (or microservices), each specialized in accomplishing a narrow range of tasks \cite{balalaie2016microservices,dragoni2017microservices}. This composability of the services facilitates independent deployment, rapid delivery, and flexible expansion of applications in many cloud architectures \cite{wang2018cloudranger}. %and enables teams to swap and re-compose components to meet new business requirements without disrupting another part of the application.
It is particularly useful in large cloud platforms such as Amazon AWS, Google Cloud, and Microsoft Azure due to their culture of team-level service ownership.

\begin{figure}[!t]
\centering
\includegraphics[width=0.95\columnwidth]{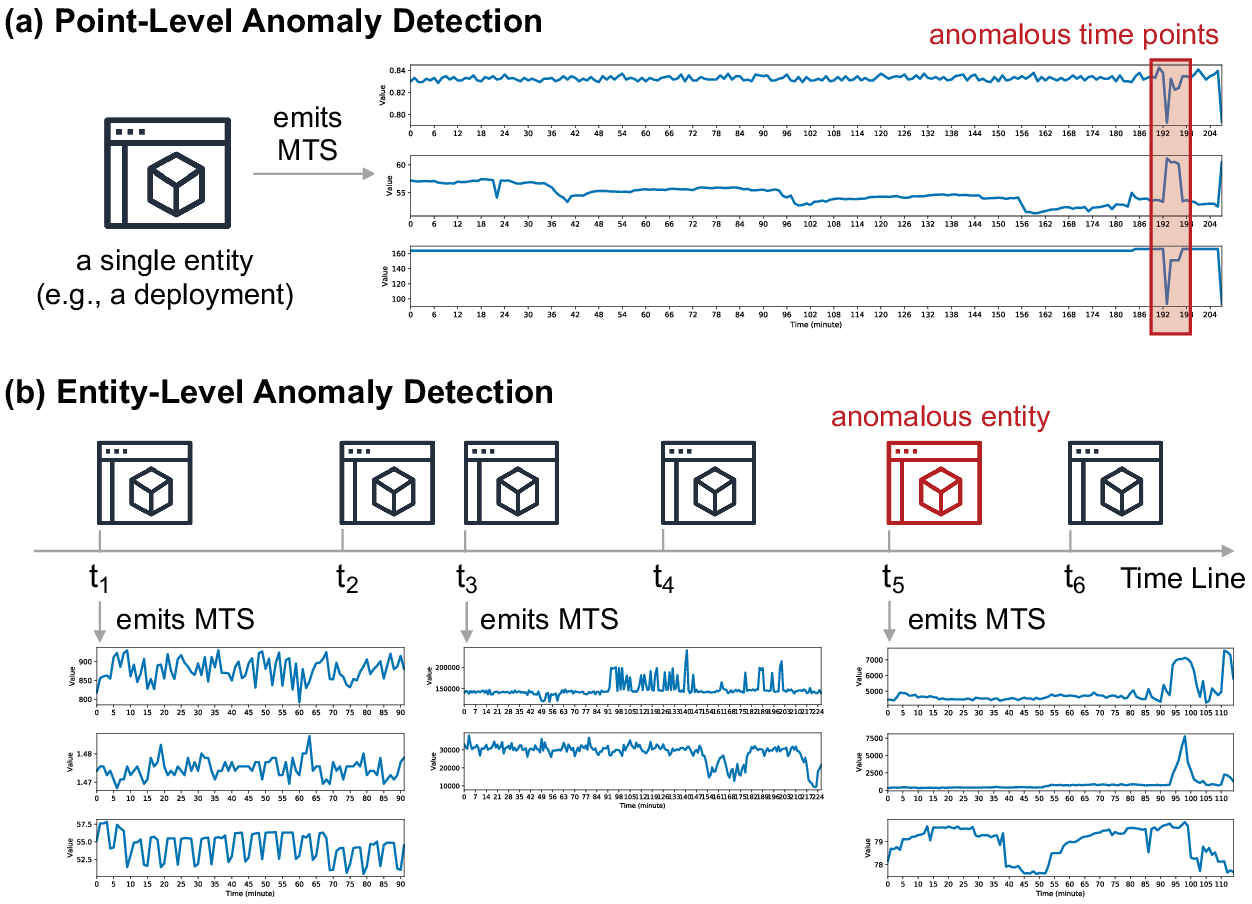}
\caption{An illustration of (a) point-level anomaly detection, and (b) entity-level anomaly detection.}\label{fig.intro}
\end{figure}

% \avkan{}{Revised first sentences below}

As the code implementing these services is regularly updated, whether to add new functionality %improve security,
or improve performance, faults may be introduced. Faulty changes can degrade the target service's performance or cause externally-facing outages, which directly impact the customer's experience and the company's reputation. To prevent faulty changes, code deployments should be monitored comprehensively for rapid detection of anomalous behaviors so that faulty deployments can be stopped in time and the software can be returned to its previous, safe status. This process is called {\em rollback}, which helps avoid cascading damages to a cloud application.

The existing approaches mainly detect anomalies in online systems from four perspectives, namely KPI-level (key performance indicator) \cite{liu2015opprentice,xu2018unsupervised}, log-level \cite{du2017deeplog,meng2019loganomaly}, trace-level \cite{gan2019seer,nedelkoski2019anomaly}, and entity-level \cite{su2019robust,huang2022semi}. In this paper, we focus on anomaly detection at the entity level, where an entity could be a cloud server, a container, or a deployment of services. For example, for a deployment of certain services at Amazon, a large number of metrics, such as CPU usage, memory usage, threads usage, {\em etc.}, are continuously monitored. Each metric emits time series (TS) and all of the metrics are aggregated into multivariate time series (MTS) as illustrated in Fig. \ref{fig.intro}(b). As such, it is intuitive to resort to MTS anomaly detection approaches for {\em entity-level anomaly detection}.

%\avkan{}{For Figure 1, it's best if it can pictorially show improvements introduced by our method. We can discuss in person. Alternatively, we can introduce a small table in the Intro.}

Existing anomaly detection techniques on MTS are mostly designed for {\em point-level anomaly detection}, which is to identify time points that have anomalous observations from their contextual time points in a single entity \cite{schmidl2022anomaly}, such as a server, as Fig. \ref{fig.intro}(a) presents. Entity-level anomaly detection is remarkably different. It aims to detect anomalous entities in a stream of entities ({\em e.g.} deployments), where each entity emits MTS, as Fig. \ref{fig.intro} (b) presents. In particular, entity-level anomaly detection poses four unique challenges.

{\bf 1. Multiple Heterogeneous Entities.} To detect anomalous entities, a model should be trained on the MTS data across different entities for capturing the behavioral patterns of the entities. However, a model trained using the metrics in one entity is hard to be applied to other entities: (1) for the same metric, the time series values are non-comparable for different entities. For example, the normal CPU usage of deploying a computational service is higher than that of a notification service. This leads to different MTS spaces in different entities. (2) different entities may have different durations, thus their MTSs may last for different lengths. (3) different entities may have a different (sub)set of metrics, making their MTSs have different number of variates (or dimensions). As such, we seek a robust model that can be shared across heterogeneous entities with MTSs that have varying scales, lengths, and variates.

{\bf 2. Low Latency Tolerance.} In an online system, the constantly emergent entities form a stream, where a single entity may only live for a short time. For example, a deployment of service at Amazon may only last for a few minutes, and the duration is unknown at the onset. Therefore, it is infeasible to train a model per entity online. This consolidates the necessity to share a pre-trained model across entities. Also, it requires the model to adapt to all historical MTS of a new entity and perform inference with a low latency.

{\bf 3. Ambiguous Anomaly Definition.} Unlike the existing anomaly detection methods that detect unexpected changes from their contextual time points, the definition of anomalies in an industrial environment is more ambiguous. A change in time series may not indicate anomaly, but a normal launch of a new deployment. Instead, the definition of anomaly is that a service cannot work normally, which necessitates supervised signals from domain experts.% The supervision can guide the model to learn the complicated patterns of many correlated variates that lead to the anomalous behaviors of services.

%\avkan{}{Add a clarifying sentence at the beginning. E.g., ``Labelling deployments for our task is expensive and time consuming. Even when we obtain expert labels, they are typically at the level of the entity, i.e., without explicitly indicating anomalous time points.''}

{\bf 4. Limited Supervision.} The entity-level labels are not point-wise, {\em i.e.}, it does not locate the time points when anomalies occur, but only indicates whether the entity is anomalous by the end of its duration. Because the existing (semi-)supervised methods require point-level labels, they cannot be trained for solving our problem. Moreover, even for the entity-level labels, domain experts may make mistakes. Given the high human cost of labeling, the challenge is then to build models with few, noisy entity labels.

To address these challenges, we propose a Semi-supervised Hybrid \underline{M}odel for \underline{E}ntity-\underline{L}evel \underline{O}nline \underline{D}etection of anomal\underline{Y} (\method). \method\ is a system used in production monitoring several million deployments every month. \method\ consists of two major components, namely an online feature extractor (\fe) and a semi-supervised anomaly detection (\ad) module. \fe\ embeds the MTSs of different entities to the same, comparable features space of fixed dimension, where the features are computed dynamically and incrementally for varying MTS lengths (Challenge 1). Because feature extraction takes inference time, we design \fe\ with efficient initialization and updating capability (Challenge 2). Based on the extracted features, \ad\ leverages supervised signals for learning anomalous patterns (Challenge 3). \ad\ is a hybrid model with two sub-modules, a supervised ensemble model that is robust to noisy features and labels (Challenge 1, 4), and a semi-supervised deep one-class model that can leverage the large amount of unlabeled data for complementing the limited supervision (Challenge 4). Two strategies to combine the outputs of the two sub-modules, {\em i.e.}, an ensemble strategy and a sequential strategy, were introduced. Our contributions can be summarized as follows.
\begin{itemize}
\item We investigate a new entity-level anomaly detection problem, which is motivated by real applications in cloud native systems. Its unique challenges can not be directly addressed by existing MTS anomaly detection approaches.
\item We propose \method, a novel robust semi-supervised framework for online anomaly detection in streaming entities. It resolves the ambiguous definition of anomalies via limited labels, and leverages the vast amount of unlabeled data for enhancing its robustness and performance.
\item We evaluate \method\ using the data of 30K+ deployments with 1.2M+ time series from Amazon, and compare it with the state-of-the-art (SOTA) approaches. The results demonstrate \method\ significantly outperforms the baseline methods, with up to 56.5\% relative improvement on F1 score.
\item We deploy \method\ as a core component of an AutoRollback system on the deployments of services at Amazon, and evaluate customer experience of the enhanced system.
\end{itemize}

% Section~\ref{s:background} discusses the non-trivial structure of the data our method operates on. We then present different components of \method{} in Section~\ref{s:phoenix}. This is followed by experimental results in Section~\ref{s:experiments}. Related work is summarized in Section~\ref{s:related_w} followed by conclusion in Section~\ref{s:conclusion}.

\section{Preliminary}\label{sec.preliminary}

%\subsection{Deployments As Entities}
In this work, we consider real-time monitoring of deployment entities. A deployment is a process of updating software packages, configuration, environment variables, {\em etc.} %which can influence one or more services.% Code changes are typically deployed progressively on a fraction the instances running the target services at a time, in order to avoid causing a service interruption. Once started, a deployment will take some unknown length of time to complete ({\em e.g.}, from 1 min to 1 hour).
Each deployment affects one or more services, each service has multiple metrics, and %each metrics emits a univarite time series. 
each metric has a univariate time series. Therefore, each deployment has multiple univariate time series, each of which has a unique label (\texttt{service}, \texttt{metric}). The set of possible metrics is fixed, but there could be unlimited number of possible services. Moreover, a service may be monitored with a subset of the metrics. Therefore, each deployment is associated with a variable number of univariate time series. Fig. \ref{fig.intro}(b) illustrates the time series of three deployments.% Next, we will formalize the problem.

\subsection{Problem Statement}\label{sec.prob}

Suppose the collection of entities is $\mathcal{X} = \{\mathcal{X}_{i}\}_{i=1}^{N}$. Each entity ({\em e.g.}, deployment) $\mathcal{X}_{i}$ has a historical multivariate time series $\mat{X}_{i} = [\mat{x}_{i,1}, ..., \mat{x}_{i,n_{i}}]$, where $\mat{x}_{i,j}=[x_{i,j}^{1}, ..., x_{i,j}^{T}] \in \mathbb{R}^{T}$ ($1 \le j \le n_{i}$) is the $j$-th univariate time series associated with a unique label ($\texttt{service}_{j}$, $\texttt{metrics}_{j}$) in an observation window of size $T$. Due to the variability of \texttt{service} and the different subsets of \texttt{metrics}, the number $n_{i}$ could be different for different $\mathcal{X}_{i}$. In this paper, we also use $\mat{x}_{i}^{t}=[x_{i,1}^{t}, ..., x_{i,n_{i}}^{t}]$ to denote an observation of the $n_{i}$ variates of $\mathcal{X}_{i}$ at time step $t$, and use $\mat{X}_{i}^{t-w:t}=[\mat{x}_{i,1}^{t-w:t}, ..., \mat{x}_{i,n_{i}}^{t-w:t}]$ to denote a sequence of observations from time $t - w$ to $t$.

To resolve the ambiguous anomaly definition, partial labels are available. Formally, let $\mathcal{X} = \{\mathcal{X}^{u}, \mathcal{X}^{l}\}$, $\mathcal{X}^{u}$ be the subset of unlabeled entities, and $\mathcal{X}^{l}$ be the labeled subset with binary labels $\mat{y} \in \{0, 1\}^{N_{l}}$, where $N_{l}=|\mathcal{X}^{l}|$ and $\mathcal{X}^{u}\cap\mathcal{X}^{l}=\varnothing$. Label $\mat{y}_{i}=1$ indicates the $i$-th entity in $\mathcal{X}^{l}$ is anomalous; $\mat{y}_{i}=0$ otherwise.

The {\em entity-level anomaly detection} problem is to train a model $f: \mathcal{X} \rightarrow \mathbb{R}$, such that given a new entity $\mathcal{X}_{\text{i}}$ with its historical MTS $\mat{X}_{i} \in \mathbb{R}^{T \times n_{i}}$, the output $f(\mat{x}_{i}^{t}|\mat{X}_{i})$ represents the anomalous score of the observation $\mat{x}_{i}^{t} \in \mathbb{R}^{n_{i}}$ at time step $t$ ($t > T$).

% \avkan{}{You need to explain where the historical time series is from and what is the significance of this. E.g., mention this in Sect. 2.1. Also, to make a decision at t>T, don't you also need points between T and t?}

\vspace{0.1cm}

\noindent{\bf Remark.} It is noteworthy that although the model $f$ should check anomalies at new time points for timely detection for at inference time, this problem is different from point-level anomaly detection because (1) the model $f$ is shared across entities with different MTSs; and (2) at training time, the label $\mat{y}$ only marks anomalous entities, without marking anomalous time points, which are required by the existing (semi-)supervised methods \cite{jiang2021semi,schmidl2022anomaly,huang2022semi,chen2023semisupervised}.

% \avkan{}{Consider re-phrasing or clarifying the first argument. In most of machine learning settings, the model is shared over a space of data. How is our setup different?}

% \avkan{}{Somewhere in this sub-section, consider reminding and re-iterating the variable nature of metrics. In particular, the first time series dimension for one deployment does not have the same interpretation as the first dimension of another deployment.}

%Accordingly, a false negative, is a faulty deployment that was allowed to complete. Such a deployment can have severe consequences, and hence the cost of a false negative error is high.

%A false positive, is a ``benign'' deployment that was rolled back. Such errors means unnecessarily preventing developers from introducing changes in services. At best this will result in lost time, as the developers will need to restart the deployment. The consequences can be more severe too. Developers can decide to opt-out of the automatic rollback feature which increases the risk of faulty deployments being successful in the future. False positives can also slow the process of deploying critical code changes, for example security patches in response to the discovery of a zero-day vulnerability.  
\section{Proposed System Overview}\label{sec.system}

% \avkan{}{This system is an important part of our contribution. This system has been created as a part of our project. Accordingly, (i) from section header and text it should be clear that the system is a part of our proposed solution, (ii) consider mentioning it in the list of contributions.}

% \avkan{}{We need to make it very clear that the method is in production. We need to mention it several times.}

\begin{figure*}[!t]
\centering
\includegraphics[width=1.8\columnwidth]{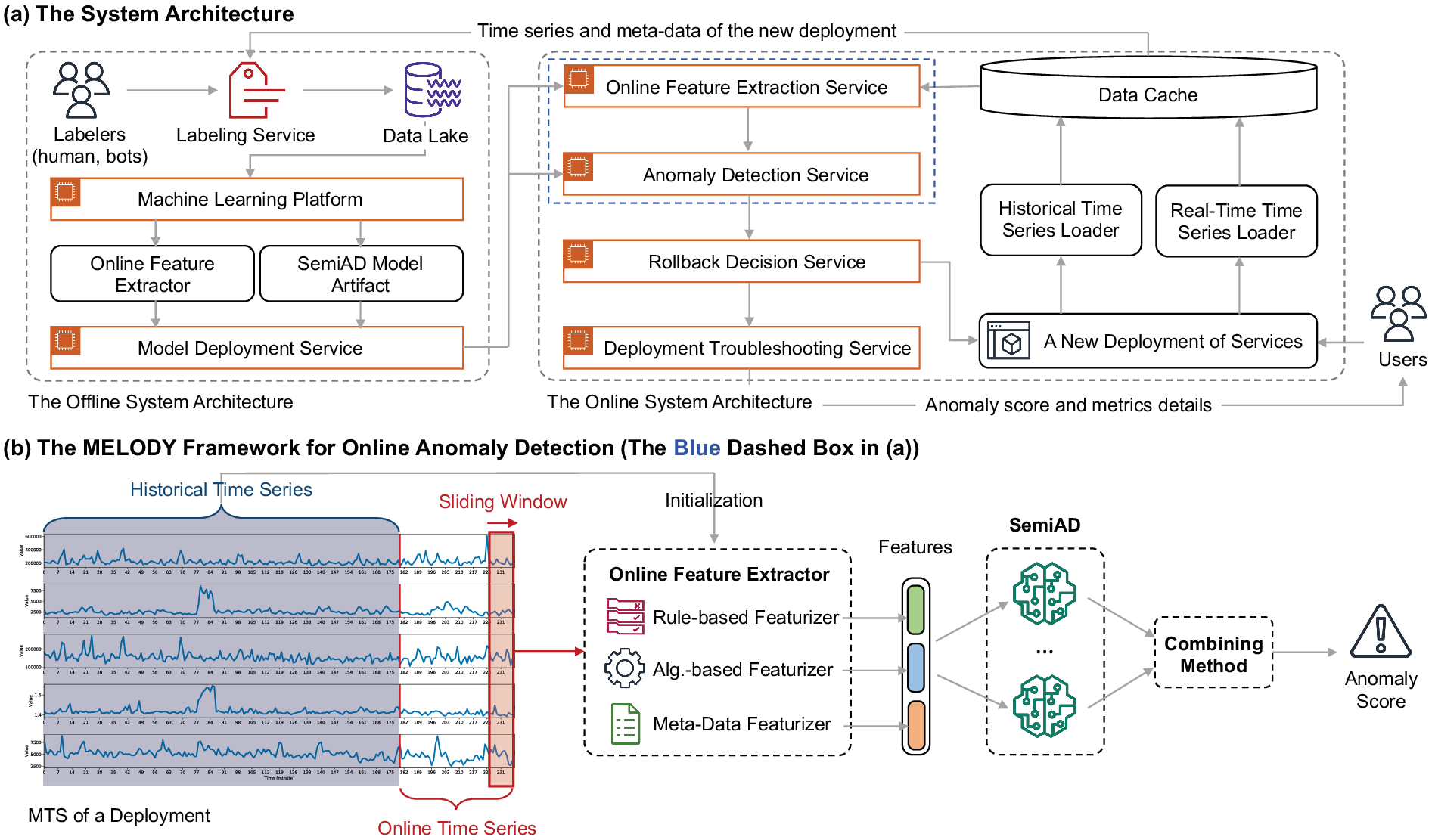}
\caption{An illustration of (a) the system architecture, and (b) the inference process of the \method\ framework.}\label{fig.system}
\end{figure*}

Fig. \ref{fig.system}(a) illustrates the system architecture for running our \method\ model for detecting anomalous deployments of services.% Fig. \ref{fig.system}(b) illustrates the details of the \method\ framework for online anomaly detection (the blue dashed box in Figure \ref{fig.system}(a)).

\subsection{The System Architecture}\label{sec.system_architecture}

% \avkan{}{Clarify that in 2.a left, we use historical data from deployments that have already happened.}

% \avkan{}{There might be concerns that we are revealing too much information. I suggest simplifying the figure as follows. Remove word ``service''. Remove blocks ``data lake'', ``model deployment'', and ``deployment troubleshooting''. Remove ``... of Services'' from ``A New Deployment ... ''. Also rename ``Customers'' to ``Developers''.}

The offline system in Fig. \ref{fig.system}(a, left) consists of three key components: (1) the data labeling service for importing data from products to experimental environment, (2) the machine learning platform for developing and training the \method\ model, and (3) the model deployment service for deploying the \fe\ and \ad\ artifacts of \method\ model to the online system.

In particular, the labeling service imports MTS and meta-data ({\em e.g.}, config profiles) of each deployment to the datalake. It provides two ways of labeling the deployments. The first is a labeling UI which visualizes the MTS %(with the (\texttt{service}, \texttt{metrics}) label on each variate)
of the deployment to be labeled. The labelers (human domain experts) use the UI to assign a score at the scale of -1 to 3 for deployments in $\mathcal{X}^{l}$: -1 means the labeler is unsure, 0 (or 1) indicates a normal (or likely normal) deployment, 3 (or 2) indicates an abnormal (or likely abnormal) deployment.

Because human labeling is costly, the second way is to automatically label normal deployments, which is referred as Bot in Fig. \ref{fig.system}(a). Bot applies a set of expert-defined rules ({\em e.g.}, whether a deployment has passed some safety checks) on the unlabeled deployments in $\mathcal{X}^{u}$. The passed deployments are automatically labeled as ``normal''. These auto-labeled ``normal'' deployments may include noise ({\em i.e.}, anomalies), but the amount should be small because anomalies are usually rare, as is the case in most anomaly detection tasks \cite{schmidl2022anomaly,han2022adbench}. This auto-labeled normal set is valuable for unsupervised or semi-supervised methods ({\em e.g.}, one-class models) to learn normal patterns.

Fig. \ref{fig.system}(a, right) is the online system. Once users launch a new deployment, the historical MTS preceding the deployment are loaded into a data cache for initializing the \fe\ module (as detailed in Sec. \ref{sec.ofe}). Then, the most recent $w$ time steps of the real-time observations are sent to the \fe\ service by the data cache, where $w$ is a window size. The \fe\ service transforms the data to features and sends them to the anomaly detection service to calculate anomalous score at the current time step, which, together with a threshold and some safety checking rules, are used by the rollback decision service to determine whether to rollback the deployment. If rolled back, the deployment trouble shooting service will display the time-wise anomaly scores and the details of relevant metrics to the users for debugging through a web interface.

In summary, the architecture enables continuous labeling of the data, and scheduled training cycle for automatic model updates with new data. In the next, we will focus on the design of the \method\ framework (the blue dashed box in Fig. \ref{fig.system}(a)).

%\subsection{The Online Anomaly Detection}

%Figure \ref{fig.system}(b) presents the details of the \method\ framework for online anomaly detection. It consists of the \fe\ and \ad\ modules. As described by the Challenge 1 in Section \ref{sec.intro}, directly using $\mat{X}_{i}^{t-w:t}$ as the input to \ad\ is infeasible because of the heterogeneity of the entities. Thus we design \fe\ to transform the different MTS’s of different entities to the same, comparable feature space of fixed dimension. \fe\ contains three parts: the rule-based features, anomaly features, and meta-features, which are initialized efficiently using the historical data. The details will be introduced in Section ???. With the extracted features, we design \ad\ to be a hybrid model consisting of a supervised ensemble model and a semi-supervised one-class model, with two combining strategies for generating the anomaly score. The \ad\ module will be introduced in Section ???.

\section{The Proposed \method\ framework}\label{sec.method}

Fig. \ref{fig.system}(b) illustrates the \method\ framework for online anomaly detection. It consists of \fe\ and \ad\ modules. The \fe\ transforms different MTSs of different entities to a comparable feature space. The \ad\ is a hybrid model that consumes the features for generating an anomaly score with a combining strategy.

%In this section, we introduce the proposed semi-supervised hybrid model for anomaly detection (\method) framework.

\subsection{Online Feature Extraction (\fe)}\label{sec.ofe}

As described by the Challenge 1 in Sec. \ref{sec.intro}, the MTS of heterogeneous entities pose three issues that prevent a model from being trained across entities. The \fe\ aims to address them.

\subsubsection{Featurizer}\label{sec.featurizer}

%\avkan{}{I think you need one or two sentences to remind why time series are non-comparable and what do we mean by heterogeneity.}

First, to address the non-comparable time series values of different entities, we transform the raw values of individual univariate time series to their anomalous degrees. Specifically, for the $i$-th deployment $\mathcal{X}_{i}$, its $j$-th variate's observation at time step $t$, {\em i.e.}, $x_{i,j}^{t}$ ($1 \le j \le n_{i}$), is transformed to a score $s_{i,j}^{t}$ representing the deviation of $x_{i,j}^{t}$ from the normal history $\mat{x}_{i,j}=[x_{i,j}^{1}, ..., x_{i,j}^{T}]$ for any $t > T$. The more $x_{i,j}^{t}$ deviates from $\mat{x}_{i,j}$, {\em i.e.}, the higher the score $s_{i,j}^{t}$ is, the more anomalous $x_{i,j}^{t}$ is. It is noteworthy that $s_{i,j}^{t}$ defines the change of $x_{i,j}^{t}$ relative to its history $\mat{x}_{i,j}$ for any univariate time series consistently, thus it is a comparable score across entities.

% \avkan{}{From your description, it seems that the score only depends on the value at t and the historical portion. Is that correct? What about value at (t-1), ..., (t-w)?}

As illustrated in Fig. \ref{fig.system}(b), the \fe\ has three {\em Featurizers} to generate score $s_{i,j}^{t}$. Each Featurizer is a {\em general class}. Given a concrete deployment $\mathcal{X}_{i}$, each Featurizer first initializes a set of {\em instances}, one per variate using its history, for some (or all) of the variates, {\em e.g.}, an instance is initialized using the history $\mat{x}_{i,j}$ for variate $j$.

% \avkan{}{What is an ``instance''? Consider calling it a ``state'', ``context variable'', ``historical summary'', or something like that.}

Then the $j$-th instance is used to transform $x_{i,j}^{t}$ to $s_{i,j}^{t}$ in an online manner. Because of the Low Latency Tolerance (Challenge 2 in Sec. \ref{sec.intro}), we design three Featurizers that initialize instances efficiently:

% \avkan{}{Somewhere in this section, consider adding an equation about generic state update. E.g., s(0) = f(x(1),...,x(T)), s(t) = g(s(t-1), x(t))}

\vspace{0.1cm}

\noindent{\bf Rule-based Featurizer.} We integrate the prior knowledge of domain experts into our model by three types of rule-based features: (1) statistics based features (SbF), (2) threshold based features (TbF), and (3) count based features (CbF).

At the initialization stage, SbF learns the mean $\mu$ and standard deviation $\sigma$ of all $T$ values in the history $\mat{x}_{i,j}$, and defines a threshold $\tau = \mu + \alpha * \sigma$, where $\alpha$ is a multiplier. At the inference stage, it sets $s_{i,j}^{t}=1$ at time step $t$ if it observes $w_{s}$ continuous values $x_{i,j}^{t-w_{s}+1}$, ..., $x_{i,j}^{t}$ above $\tau$; and sets $s_{i,j}^{t}=0$ otherwise. By defining different $\alpha$ and $w_{s}$ for the metrics of interest, SbF has a total of 11 instances.

TbF is similar to SbF except that TbF uses a predefined threshold $\tau$ directly without resorting to $\mu$ and $\sigma$. By defining $\tau$ and $w_{s}$, TbF has 7 possible instances. CbF is used to count the number of continuous missing observations, which can indicate anomalies on certain metrics. CbF's initialization sets up a sliding window of size $w_{c}$. At the inference stage, CbF sets $s_{i,j}^{t}=1$ if the window ending at time step $t$ is full of missing observations; and sets $s_{i,j}^{t}=0$ otherwise. CbF has 1 possible instance.

It is noteworthy that all of the SbF, TbF, CbF Featurizers have efficient initialization. In total, rule-based featurizer has 19 possible instances for the metrics that the rules monitor.

% \avkan{}{Are we updating mean and st. dev. as new observations arrive? If not, it might be worth adding a sentence explaining why. E.g., are we treating historical time series as ``normal data''?}

\vspace{0.1cm}

\noindent{\bf Algorithm-based Featurizer.} We also integrate two efficient point-level anomaly detection algorithms on univariate time series for scoring $s_{i,j}^{t}$: subsequence-based nearest neighbor (SubNN) \cite{schmidl2022anomaly} and median forecast (MD) \cite{basu2007automatic}.

SubNN is a distance based scorer. At the initialization stage, SubNN segments the history $\mat{x}_{i,j}$ into subsequences with window size $w_{n}$ and stride 1, forming a set $\mathcal{S}$ of length-$w_{n}$ subsequences, and sets up a sliding window of size $w_{n}$ for the inference stage. At the inference stage, SubNN sets $s_{i,j}^{t}$ as the distance between the sliding window ending at time step $t$ and its nearest neighbor in $\mathcal{S}$. MD is an efficient forecasting approach. At the initialization stage, it only sets up a sliding window of size $w_{m}$. At the inference stage, based on the sliding window at $t-1$, it calculates $m_{\text{obs}}=\text{median}([x_{i,j}^{t-w_{m}}, ..., x_{i,j}^{t-1}])$, and $m_{\text{dif}}=\text{median}([x_{i,j}^{t-w_{m}+1} - x_{i,j}^{t-w_{m}}, ..., x_{i,j}^{t-1} - x_{i,j}^{t-2}])$, and forecasts the subsequent value $\hat{x}_{i,j}^{t}=m_{\text{obs}} + (w_{m}/2)*m_{\text{dif}}$. We adapted it for scoring anomalies by setting $s_{i,j}^{t}$ as the deviation of $x_{i,j}^{t}$ from $\hat{x}_{i,j}^{t}$, {\em i.e.}, $s_{i,j}^{t} = |x_{i,j}^{t} - \hat{x}_{i,j}^{t}|$.

The algorithm-based featurizers are applicable to any metric. There are 22 possible metrics in total, thus the SubNN and MD featurizers have 44 instances.

\vspace{0.1cm}

\noindent{\bf Meta-Data Featurizer.} For entity-level anomaly detection, we append some static meta-data features of a deployment to the time-wise features emitted by the rule-based and algorithm-based featurizers, as illustrated by Fig. \ref{fig.system}(b). There are 8 meta-data features pertaining to the configurations of each deployment, which are useful as anomalous pattern may vary with configurations.% A list of these features can be found in Appendix ?.

% \avkan{}{We might not be able to reveal details.}

\subsubsection{Time Pooling Layer}\label{sec.time_pooling_layer}

The second issue in Challenge 1 (Sec. \ref{sec.intro}) is the variable duration of different deployments. Using any instance of the rule-based and algorithm-based featurizer, we can obtain a feature $s_{i,j}^{t}$ at time step $t$. However, as described in Sec. \ref{sec.preliminary}, the labels $\mat{y}$ of the training deployments in $\mathcal{X}^{l}$ are at the entity-level, where $\mat{y}_{i}=1$ only indicates the deployment $\mathcal{X}_{i}$ is anomalous before it ended, without marking any anomalous time points. Thus we cannot train a model on the feature $s_{i,j}^{t}$ at a specific time point $t$ using $\mat{y}_{i}$. To address it, we seek for an entity-level method that (1) can be updated efficiently to track dynamic features, and (2) is invariant to the variable duration of deployments.

To this end, we designed a time pooling layer with two pooling methods up to the current time point $t$:
\begin{equation}\label{eq.pooling}
\begin{aligned}
\hat{s}_{i,j}^{t} = \text{MaxPool}([s_{i,j}^{1}, ..., s_{i,j}^{t}]),~~~\bar{s}_{i,j}^{t} = \text{MeanPool}([s_{i,j}^{1}, ..., s_{i,j}^{t}])
\end{aligned}
\end{equation}
where the $\hat{s}_{i,j}^{t}$ records the most anomalous status up to time point $t$, and $\bar{s}_{i,j}^{t}$ records the accumulative anomalous status.

The pooling methods are invariant to variable sequence lengths. % \cite{zaheer2017deep},
For model training, Eq.~\eqref{eq.pooling} generates a feature per training deployment by pooling up to the end of its length. Also, because both MaxPool and MeanPool can be updated incrementally in constant time, % \cite{gomes2019machine},
we can use Eq.~\eqref{eq.pooling} to dynamically track salient features of new deployments up to any time point for online inference.

% \avkan{}{I don't think the above paragraph needs citations}

% \avkan{}{Did we also consider the feature itself? E.g., instead of applying pooling just use the latest feature itself s[i,j,t] = s[i,j,t]}

\subsubsection{Feature Aggregator}\label{sec.aggregator}

% \avkan{}{This section is difficult to understand. The concept is tricky. I'm not sure what's the best way to explain it. One suggestion is to introduce another index, and say z[i,k,t] = MaxPool(s[i,j,t] |  metricID(s[i,j]) == k). Here j index time series, and we can have a variable number of them. k index a fixed set of metrics. Another suggestion is to have a running example in appendix.}

So far, using any of the featurizer instances, we can generate features $[\hat{s}_{i,j}^{t}, \bar{s}_{i,j}^{t}]$ for each univariate time series of $\mathcal{X}_{i}$, which has a unique label ($\texttt{service}_{j}$, $\texttt{metrics}_{j}$) (Sec. \ref{sec.prob}). The third issue in Challenge 1 (Sec. \ref{sec.intro}) implies the number of variates $n_{i}$ is different for different $\mathcal{X}_{i}$, leading to different feature dimensions for different $\mathcal{X}_{i}$. There are two reasons for the different $n_{i}$: (1) the deployments can have different subsets of metrics, and (2) multiple services may be monitored for the same metric, generating multiple univariate time series on the same metric.

To address this challenge, and embed different deployments in the same feature space, we propose a feature aggregator. After we obtain the feature $[\hat{s}_{i,j}^{t}, \bar{s}_{i,j}^{t}]$ from Eq.~\eqref{eq.pooling} for each univariate time series, we aggregates the features over different services for the same metric. %Suppose the set of total 22 metrics is $\mathcal{M}=\{m_{k}\}_{k=1}^{22}$.
Suppose $m_{k}$ is the $k$-th metric. Taking $\hat{s}_{i,j}^{t}$ as an example, we perform an aggregation for $m_{k}$
\begin{equation}\label{eq.agg}
\begin{aligned}
\hat{z}_{i,k}^{t} = \text{MaxPool}(\{\hat{s}_{i,j}^{t}|\texttt{metrics}_{j}=m_{k}, 1 \le j \le n_{i}\})
\end{aligned}
\end{equation}
where MaxPool is used because we want to keep the most salient anomalous feature from different services.

Similarly, we can obtain $\bar{z}_{i,j}^{t}$ from $\bar{s}_{i,j}^{t}$, and this step addresses the issue (2). To address issue (1), if a deployment misses a specific metric value $m_{k}$ and Eq.~\eqref{eq.agg} cannot be applied for $m_{k}$, we perform mean-based imputation on $\hat{z}_{i,k}^{t}$ using the training deployments in $\mathcal{X}$. Thus we align all deployments to the same set of 134 features (63 $\hat{z}_{i,k}^{t}$, 63 $\bar{z}_{i,k}^{t}$ from the 63 featurizer instances and 8 meta-data features in Sec. \ref{sec.featurizer}). Then each deployment $\mathcal{X}_{i}$ is represented by a vector $\mat{z}_{i} \in \mathbb{R}^{d}$ of fixed dimension $d=134$.%To address the potentially noisy imputation on missing metrics, in next section, we propose a semi-supervised hybrid model on $\{\mat{z}_{i}\}_{i=1}^{N}$ for selecting valid features and learning meaningful embeddings using the supervision signals.

To address noisy imputation and model the correlation of variates in MTS, next, we propose a semi-supervised model on $\{\mat{z}_{i}\}_{i=1}^{N}$ for learning meaningful embeddings using supervision signals.

\subsection{Semi-Supervised Anomaly Detection}

Taking features $\mat{z}_{i} \in \mathbb{R}^{d}$, \ad\ aims to train a detector $f: \mathbb{R}^{d} \rightarrow \mathbb{R}$, and uses it to emit the final anomaly score $\hat{y}_{i}$ for every deployment $\mathcal{X}_{i}$. To be robust to the imputed values in $\mat{z}_{i}$, and to address Challenge 3 (ambiguous anomaly definition) and 4 (limited supervision) in Sec. \ref{sec.intro}, we propose a hybrid model comprising a semi-supervised one-class model and a supervised ensemble model.

% \avkan{}{should it map to [0,1]?}
% \avkan{}{remind what the challenges are, e.g., ``to address Challenge 3 (ambiguity in defining anomalies) and 4 (limited amount of labels)''}

\subsubsection{Semi-Supervised Deep One-Class Model}\label{sec.semioc}

A one-class model is an unsupervised anomaly detector that is trained on samples of a single, typically normal, class. It is used to predict whether a testing sample belongs to this class or not. One prominent example is kernel-based method, such as One-Class SVM (OC-SVM) \cite{scholkopf2001estimating} and Support Vector Data Description (SVDD) \cite{tax2004support}, which aims to maximize the margin between normal samples and others. For example, SVDD aims to find the smallest hypersphere with a center vector $\mat{c}$ and a radius $R>0$ to enclose the majority of the (normal) samples in the feature space.
\begin{equation}\label{eq.svdd}
\begin{aligned}
&\min_{\mat{c}, R, \boldsymbol{\xi}}~~R^{2} + \frac{1}{\nu N}\sum_{i=1}^{N}\xi_{i}\\
&\text{s.t.}~~\|\phi(\mat{z}_{i}) - \mat{c}\|^{2} \le R^{2} + \xi_{i},~~\xi_{i} \ge 0,~~\forall i = 1, ..., N
\end{aligned}
\end{equation}
where $\phi(\mat{z}_{i})$ is a kernel function on input feature $\mat{z}_{i}$, $\xi_{i}$ is a slack variable to allow soft boundary, and $\nu \in (0, 1]$ is a hyperparameter to control the trade-off between the volume of the sphere and the penalties on $\xi_{i}$. Once $\mat{c}$ and $R$ are determined by solving the dual form of the primal problem in Eq.~\eqref{eq.svdd}, samples that are outside the sphere, {\em i.e.}, $\|\phi(\mat{z}_{i}) - \mat{c}\|^{2} > R^{2}$, are deemed anomalies.

Recently, DeepSVDD was introduced in \cite{ruff2018deep}. Compared to the kernel-based methods, it is more robust to the noises from feature engineering, and more scalable to large datasets. It replaces $\phi(\cdot)$ by a neural network $\phi(\cdot; \boldsymbol{\theta}): \mathbb{R}^{d} \rightarrow \mathbb{R}^{d_{e}}$ with parameter $\boldsymbol{\theta}$, where $d_{e}$ is the dimension of the embedding space. The objective of DeepSVDD is to train $\boldsymbol{\theta}$ for learning a transformation $\phi(\cdot; \boldsymbol{\theta})$ that minimizes the volume of a hypersphere centered on a predetermined $\mat{c}$.
\begin{equation}
\begin{aligned}
\min_{\boldsymbol{\theta}}\frac{1}{N}\sum_{i=1}^{N}D\big(\phi(\mat{z}_{i}; \boldsymbol{\theta}), \mat{c}\big) + \lambda\|\boldsymbol{\theta}\|_{F}^{2}
\end{aligned}
\end{equation}
where $D(\cdot, \cdot)$ is a distance function, such as Euclidean or Hamming distance, and $\lambda$ is the hyperparameter for weight decay.

%Penalizing the distance of embeddings to the center $\mat{c}$ forces the network to extract those common factors of variation which are most stable within the dataset. Then normal samples tend to be close to the center, while anomalies are further away.

Although DeepSVDD can learn embeddings that are less sensitive to the noises in $\mat{z}_{i}$, %from our Feature Aggregator (Sec. \ref{sec.aggregator}),
it can not harness the supervised signals for learning embeddings that resolve the ambiguous anomaly definition (Challenge 3 in Sec. \ref{sec.intro}). Recent works also indicated even with a small amount of labels, semi-supervised methods could outperform unsupervised methods significantly on anomaly detection \cite{han2022adbench}.

Therefore, we introduce our Semi-Supervised Deep One-Class Model (SemiDOC). Our goal is to use the small amount of labeled anomalies to tighten the boundary of the hypersphere. To this end, we propose a negative sampling based model trained in batch-wise. In each batch, we randomly sample $B$ normal samples from the labeled set $\mathcal{X}^{l}$ or the unlabeled set $\mathcal{X}^{u}$ (which is auto-labeled as normal as described in Sec. \ref{sec.system_architecture}) as the queries $\{\mat{z}_{i}^{q}\}_{i=1}^{B}$. For each query $\mat{z}_{i}^{q}$, we sample an anomaly from $\mathcal{X}^{l}$ as its negative sample, and form a tuple $(\mat{z}_{i}^{q}, \mat{z}_{i}^{n})$. Then each batch is a set of $B$ tuples $\{(\mat{z}_{i}^{q}, \mat{z}_{i}^{n})\}_{i=1}^{B}$, and the learning objective is
\begin{equation}\label{eq.semioc}
\begin{aligned}
\min_{\boldsymbol{\theta}}\frac{1}{B}\sum_{i=1}^{B}\Big(D\big(\phi(\mat{z}_{i}^{q}; \boldsymbol{\theta}), \mat{c}\big) + \ell(\mat{z}_{i}^{q}, \mat{z}_{i}^{n}; \boldsymbol{\theta})\Big) + \lambda\|\boldsymbol{\theta}\|_{F}^{2}
\end{aligned}
\end{equation}
where
\begin{equation}\label{eq.hinge}
\begin{aligned}
\ell(\mat{z}_{i}^{q}, \mat{z}_{i}^{n}; \boldsymbol{\theta}) = \max{\Big(\delta - D\big(\phi(\mat{z}_{i}^{q}; \boldsymbol{\theta}), \phi(\mat{z}_{i}^{n}; \boldsymbol{\theta})\big), 0\Big)}
\end{aligned}
\end{equation}
is a hinge loss to maximize the distance between the embeddings of $\mat{z}_{i}^{q}$ and $\mat{z}_{i}^{n}$, and $\delta$ is a threshold to avoid arbitrarily large distance values in the loss function, for training stability.

After training the model, SemiDOC uses the following function to infer anomaly score of a new sample $\mat{z}_{\text{new}}$.
\begin{equation}\label{eq.semiocscore}
\begin{aligned}
\text{AnomalyScore}(\mat{z}_{\text{new}}; \boldsymbol{\theta}) = \text{Clip}{\bigg(\frac{D\big(\phi(\mat{z}_{\text{new}}; \boldsymbol{\theta}), \mat{c}\big)}{R}, 0, 1\bigg)}
\end{aligned}
\end{equation}
where $R=\max_{\mat{z}_{i} \in \mathcal{Z}_{\text{normal}}}D\big(\phi(\mat{z}_{i}; \boldsymbol{\theta}), \mat{c}\big)$ is the maximal radius of the normal embeddings in set $\mathcal{Z}_{\text{normal}}$ in the training set, {\em i.e.}, the learned hypersphere, and the Clip function is used to prevent extreme values from dominating the anomaly scores.

\vspace{0.1cm}

\noindent{\bf Remark.} Our method is different from DeepSAD \cite{ruff2019deep}, which only regularizes the distance between the labeled samples and the center $\mat{c}$, without explicit manipulation on the boundary of normal data. In contrast, the proposed SemiDOC uses negative sampling in Eq.~\eqref{eq.hinge} to explicitly tighten the boundary of normal embeddings, facilitating the detection of hard anomalies that are close to the boundary. We empirically demonstrate the superiority of SemiDOC in Sec. \ref{sec.ablation}.

\subsubsection{Supervised Anomaly Detector}

Given the intricate anomalous patterns and the potential presence of noise in the unlabeled set $\mathcal{X}^{u}$, a semi-supervised model may be biased from learning an accurate boundary. To address it, we add a robust supervised model to provide another angle of class boundary, and combine it with SemiDOC in an ensemble in Sec. \ref{sec.hybrid}. Our empirical findings in Sec. \ref{sec.exp} consolidate the superiority of such a hybrid model.

We employ LightGBM \cite{ke2017lightgbm} as the supervised detector. LightGBM is a boosting tree-based ensemble method that has high accuracy and efficiency. As a binary classifier, it has been demonstrated as useful in anomaly detection tasks \cite{vargaftik2021rade,han2022adbench}, %It has been widely used in many winning solutions of ML competitions,
and been deployed in many production pipelines of fraud prevention systems. Similar methods such as XGBoost \cite{chen2016xgboost} and CatBoost \cite{prokhorenkova2018catboost} are also compatible with our framework. We selected LightGBM for its better efficiency and superior accuracy. By feeding a feature $\mat{z}_{i}$ to LightGBM, we use the probability to the anomalous class as the anomaly score.

\subsubsection{Hybrid Model}\label{sec.hybrid}

Ensembling was found as a powerful regularization technique for performance improvement \cite{oreshkin2019n}. We ensemble SemiDOC and LightGBM to form a hybrid model for anomaly detection. The core property of an ensemble is diversity. Thus we perform a bagging procedure \cite{breiman1996bagging} by including models (SemiDOC or LightGBM) trained with different random initializations.

As for the ensemble aggregation function, the widely used approach is taking mean of the anomaly scores from different models in the hybrid. However, as a one-class model, a high score from SemiDOC in Eq.~\eqref{eq.semiocscore} does not necessarily mean anomalies, but indicates an unknown entity is different from the normal majority of the training set. Thus taking mean of the scores of SemiDOC and LightGBM may generate a high score for unknown but normal entities, leading to more false positives.

To alleviate it, we propose a sequential approach with two steps. First, SemiDOC is used to filter normal entities with scores lower than a threshold. Second, the entities that SemiDOC is less confident ({\em i.e.}, with high scores) are sent to LightGBM for anomaly detection. %This sequential hybrid model is illustrated in Fig. \ref{}.
If there are multiple SemiDOC (or LightGBM) in the hybrid, the mean score of SemiDOCs (or LightGBMs) is used in the two steps.
%Its drawback is it may be biased to the known anomalies from the training set, leading to more false negatives. Therefore, there is a trade-off in the two aggregation methods.

In our experiments, we evaluated both mean-based and sequential models, which are named as \method-M and \method-S.

\subsection{Time Complexity Analysis}

We analyzed the time complexity of \fe\ and \ad\ in detail in Appendix \ref{sec.complexity}. In summary, the time complexity of \method\ for anomaly inference is approximately $O(T)$, which is efficient as $T$ can be fixed as a constant length of historical time series.

\section{Experiments}\label{sec.exp}

\subsection{Datasets}

\begin{table}[!t]
\caption{The statistics of dataset}\label{tab.dataset}
\vspace{-0.3cm}
\begin{center}
\begin{tabular}{lccc}\hline
{\bf Dataset} & {\bf \# entities} & {\bf \# time series} & {\bf \# anomalies} \\ \hline
Hard-Labeled set & 4,966 & 234,508 & 288 (5.8\%) \\
Soft-Labeled set & 4,966 & 234,508 & 544 (11.0\%) \\
Naive-Labeled set & 4,688 & 220,320 & 544 (11.6\%) \\
Unlabeled set & 27,590 & 1,034,711 & NA \\ \hline
\end{tabular}
\end{center}
\vspace{-0.3cm}
\end{table}

\begin{table*}[!t]
\caption{The performance on anomaly detection of the compared methods. $\uparrow$ means higher is better. $\downarrow$ means lower is better. The best and second results in the F1 column ({\em i.e.}, the overall performance metrics) are in bold and underlined, respectively.% \guinetgg{}{Could there be ways to add alternatives not using OFE ? For instance, directly mean/max pooling over the raw value of the TS ? By using General vs Ours, I feel like it diminishes the fact that OFE is still a contribution of the paper as well.}
}\label{tab.results}
\vspace{-0.3cm}
\begin{center}
\small
\begin{tabular}{ll|c|ccc|c|ccc|c|ccc}\hline
\multicolumn{2}{l}{\multirow{2}{*}{\bf Method}} & \multicolumn{4}{|c}{\bf Hard Labels} & \multicolumn{4}{|c}{\bf Soft Labels} & \multicolumn{4}{|c}{\bf Naive Labels}\\ \cline{3-14}
& & F1 $\uparrow$ & Prec. $\uparrow$ & Recall $\uparrow$ & FPR $\downarrow$ & F1 $\uparrow$ & Prec. $\uparrow$ & Recall $\uparrow$ & FPR $\downarrow$ & F1 $\uparrow$ & Prec. $\uparrow$ & Recall $\uparrow$ & FPR $\downarrow$ \\ \hline
\multirow{2}{*}{MTS} & OmniAnomaly & 0.206 & 0.126 & 0.573 & 0.241 & 0.294 & 0.193 & 0.622 & 0.317 & 0.311 & 0.209 & 0.608 & 0.296\\
& AnomalyTrans & 0.217 & 0.132 & 0.616 & 0.246 & 0.298 & 0.210 & 0.514 & 0.236 & 0.316 & 0.227 & 0.521 & 0.229\\
& TranAD & 0.207 & 0.126 & 0.573 & 0.240 & 0.297 & 0.194 & 0.629 & 0.318 & 0.309 & 0.209 & 0.594 & 0.289\\ \hline
\multirow{8}{*}{General} & \fe+DeepSVDD & 0.188 & 0.168 & 0.357 & 0.174 & 0.286 & 0.203 & 0.666 & 0.422 & 0.357 & 0.274 & 0.533 & 0.190\\
& \fe+DeepSVDD-B & 0.221 & 0.176 & 0.348 & 0.107 & 0.316 & 0.214 & 0.717 & 0.379 & 0.332 & 0.231 & 0.631 & 0.283\\
& \fe+DeepSAD & 0.337 & 0.244 & 0.555 & 0.105 & 0.422 & 0.377 & 0.497 & 0.104 & 0.439 & 0.398 & 0.494 & 0.096\\
& \fe+DeepSAD-B & 0.355 & 0.274 & 0.508 & 0.082 & 0.437 & 0.399 & 0.485 & 0.089 & 0.449 & 0.390 & 0.532 & 0.107 \\
& \fe+RF & 0.382 & 0.300 & 0.535 & 0.077 & 0.455 & 0.397 & 0.533 & 0.098 & 0.466 & 0.475 & 0.458 & 0.065\\
& \fe+RF-B & 0.362 & 0.279 & 0.521 & 0.082 & 0.456 & 0.395 & 0.540 & 0.101 & 0.470 & 0.411 & 0.552 & 0.103 \\
& \fe+LGBM & 0.397 & 0.331 & 0.502 & 0.062 & 0.491 & 0.442 & 0.553 & 0.085 & \underline{0.518} & 0.481 & 0.564 & 0.078\\
& \fe+LGBM-B & 0.399 & 0.325 & 0.522 & 0.066 & 0.490 & 0.434 & 0.563 & 0.091 & 0.515 & 0.466 & 0.576 & 0.086\\ \hline
\multirow{2}{*}{Ours} & \method-M & \underline{0.411} & 0.333 & 0.540 & 0.066 & {\bf 0.499} & 0.436 & 0.584 & 0.092 & 0.514 & 0.464 & 0.577 & 0.086\\
& \method-S & {\bf 0.432} & 0.393 & 0.485 & 0.045 & \underline{0.493} & 0.452 & 0.546 & 0.081 & {\bf 0.544} & 0.482 & 0.625 & 0.086\\ \hline
\end{tabular}
\end{center}
\vspace{-0.2cm}
\end{table*}

We sampled real data of the deployments of a variety of services from Amazon AWS between April 7, 2022 and Dec. 29, 2022. This dataset contains 4966 labeled deployments in $\mathcal{X}^{l}$ and 27590 unlabeled deployments in $\mathcal{X}^{u}$. The deployments in $\mathcal{X}^{l}$ were labeled by the labeling service as described in Sec. \ref{sec.system_architecture}. Because one deployment may be scored by multiple human judges at the scale of -1 to 3, we used three approaches to aggregate and binarize the labels: (1) Hard Labels: $\mat{y}_{i}=1$ if all labelers scored $\mathcal{X}_{i}$ as 3; $\mat{y}_{i}=0$ otherwise, (2) Soft Labels: $\mat{y}_{i}=1$ if the scores of $\mathcal{X}_{i}$ are either 2 or 3; $\mat{y}_{i}=0$ otherwise, (3) Naive Labels: the same as Soft Labels except that $\mat{y}_{i}=0$ if the scores of $\mathcal{X}_{i}$ are either 0 or 1 (-1 were excluded). For each deployment, there are 22 monitored metrics, such as Threads, CPU usage, and Memory usage, and 8 meta-data features, such as number of services and number of hosts. For each metric, we used its 2 day observations prior to the launch of deployment $\mathcal{X}_{i}$ as its history $\mat{x}_{i,j} \in \mathbb{R}^{T}$ for the \fe\ module. Because the observations were collected in every minute, $T=2880$. The lengths of different deployments after launch could be different, and the average length is 16.1 minutes. Table \ref{tab.dataset} summarizes the statistics of the dataset.

% \guinetgg{}{Another comment from AMLC, with outdated number: "\textit{It is not clear from the paper what the rate of anomalous deployments is. In table 1 it is indicated 6.4\% over the evaluation set, but the bot data is biasing this number down since all of its labels are negative}" Do we want to tackle this estimation of anomaly rate in our work or conveniently dodge it?} \guinetgg{}{It might also be worthy to add a quick sentence connecting the challenges of the problem listed extensively listed above and the fact that the accuracy metrics are not that high.}

% \avkan{}{-1 is not explained}
% \avkan{}{We might not be able to reveal details}
% \avkan{}{Table 1 will allow the public know the proportion of defective deployments at Amazon. I don't think we can disclose this. Think of a different way of summarizing this dataset.}
% \avkan{}{For unlabelled set, set number of anomalies as unknown rather than NA}

\subsection{Experimental Setup}

\subsubsection{Baselines}\label{sec.setup}

We compare \method\ with a variety of anomaly detection (AD) methods from two groups: MTS-based methods and General AD methods. For the first group, because our entity-level AD problem does not assume the availability of point-level labels on MTS, we only evaluated unsupervised SOTA methods: (1) OmniAnomaly \cite{su2019robust}, (2) AnomalyTransformer \cite{xu2021anomaly}, (3) TranAD \cite{tuli2022tranad}. We applied these methods on the MTS of each deployment individually. Following \cite{xu2021anomaly}, once an anomalous time point is detected, the deployment is considered as anomalous. For the second group, we included unsupervised method: (4) DeepSVDD \cite{ruff2018deep}, semi-supervised method: (5) DeepSAD \cite{ruff2019deep}, and supervised methods: (6) RandomForest (RF) \cite{breiman2001random}, (7) LightGBM \cite{ke2017lightgbm}, which can use the entity-level labels by applying them on the features generated by our \fe\ module. Thus they were named with prefix ``\fe+'', {\em e.g.}, \fe+LGBM. For fair comparison, the bootstrap ensemble versions of the methods in the second group were included and named with suffix ``-B'', {\em e.g.}, \fe+LGBM-B.

For our method, we considered two variants: \method-M and \method-S as discussed in Sec. \ref{sec.hybrid}. We used Hamming distance $D(\cdot, \cdot)$ in Eq.~\eqref{eq.semioc}. The hyperparameter $\delta$, {\em i.e.} the threshold in Eq.~\eqref{eq.hinge} was grid searched in \{1, 10, 100\} using validation set. By default, we used 3 LightGBMs and 3 SemiDOCs in our method, with a total 6 models. For fair comparison, each of the ensemble baseline used 6 models. In Sec. \ref{sec.parameter} and Sec. \ref{sec.ablation}, we performed hyperparameter study and ablation analysis to evaluate the design choice of \method. Its implementation details are in Appendix \ref{sec.implementation}.

%For the neural network $\phi(\cdot; \boldsymbol{\theta})$ in SemiDOC of \method, we implemented a two layer encoder with embedding dimension 128. LeakyRELU (slope 0.1) was used as the activation function in the input layers, and Sigmoid was used as the output activation. LayerNorm \cite{ba2016layer} was added in each layer. The model was optimized by Adam \cite{kingma2014adam} with learning rate 0.001, weight decay 0.0001, batch size 256, and a maximum of 500 epochs. Early stopping was employed using the validation set. We used Hamming distance $D(\cdot, \cdot)$ in Eq.~\eqref{eq.semioc}. The hyperparameter $\delta$, {\em i.e.} the threshold in Eq.~\eqref{eq.hinge} was grid searched in \{1, 10, 100\} using validation set. By default, we used 3 LightGBMs and 3 SemiDOCs in our method, with a total 6 models. For fair comparison, each of the ensemble baseline used 6 models. In Sec. \ref{sec.parameter} and Sec. \ref{sec.ablation}, we performed hyperparameter study and ablation analysis to evaluate the design choice of \method.

\subsubsection{Evaluation}

We randomly split the labeled set $\mathcal{X}^{l}$ into 60\% /20\%/20\% train/validation/test sets 5 times, and run each method on these 5 splits to evaluate the average performance. The unlabeled set $\mathcal{X}^{u}$ was only used for training semi-supervised methods as it does not have valid labels for validation and testing.

Following \cite{xu2021anomaly,huang2022semi}, we use Precision, Recall, and F1-score to evaluate the AD performance of the compared methods. Additionally, we added False Positive Rate (FPR) because in a rollback system, it is important to avoid false positives, {\em i.e.}, unnecessary rollbacks, that lead to slow deployments and poor user experience. Following \cite{huang2022semi}, for each method, we select the threshold with the best F1-score.

% \guinetgg{}{One reviewer from AMLC said: "\textit{The paper stresses the need for deployment rollback for preventing large scale issues but it is unclear why precision was chosen over recall. As author(s) states that increasing precision will help in retaining customers and preventing opt-out but this also means that it increases FN (recall decreases) thus leading to more anomalous deployments in productions. It would have been great if a stronger argument was presented as the main objective of this paper is to automatically rollback faulty deployments}" From what's above and the general story of the paper, it looks like it's less a trade-off and more a general win so the point would be less relevant. Is that your impression as well ?}

\subsection{Experimental Results}

% \avkan{}{Results are nice, but we cannot reveal the numbers. I think we can only reveal relative differences, e.g., our method is X percent points better than a baseline.}

% \avkan{}{Where are results for the first group of methods, e.g., OmniAD?}

Table \ref{tab.results} summarizes the average results of the compared methods on the test sets, from which we have several observations. First, the methods using \fe\ generally outperform MTS-based methods, indicating the unsupervised MTS-based AD methods are not applicable to entity-level anomaly detection, and it is important to align different entities to the same feature space. Our \fe\ module provides a foundation to accomplish this task. Second, for the general AD methods, their ensemble versions (with ``-B'') are overall better than the original methods ({\em e.g.}, in terms of F1), indicating the usefulness of ensembling in entity-level anomaly detection. Third, \fe+DeepSAD outperforms \fe+DeepSVDD, indicating a semi-supervised method is better than unsupervised method. Fourth, both \fe+RF and \fe+LGBM outperform \fe+DeepSAD, indicating the labeled set may be more important that the unlabeled set. Finally, both \method-M and \method-S outperform the baselines in most cases in terms of F1, {\em e.g.}, \method-S has a 7.6\% to 56.5\% relative improvement on F1 on the Hard Labels dataset. This demonstrates the effective use of both labeled and unlabeled sets by the proposed methods. In particular, \method-S is superior in precision and FPR in most cases. This attributes to the sequential design of its ensembling (Sec. \ref{sec.hybrid}), where SemiDOC can effectively filter out true normal entities, resulting in less false positives and high precision, meanwhile maintains a reasonable recall.

\subsection{More Details on Effectiveness}

\subsubsection{Parameter Analysis}\label{sec.parameter}

\begin{figure}[!t]
\centering
\includegraphics[width=0.97\columnwidth]{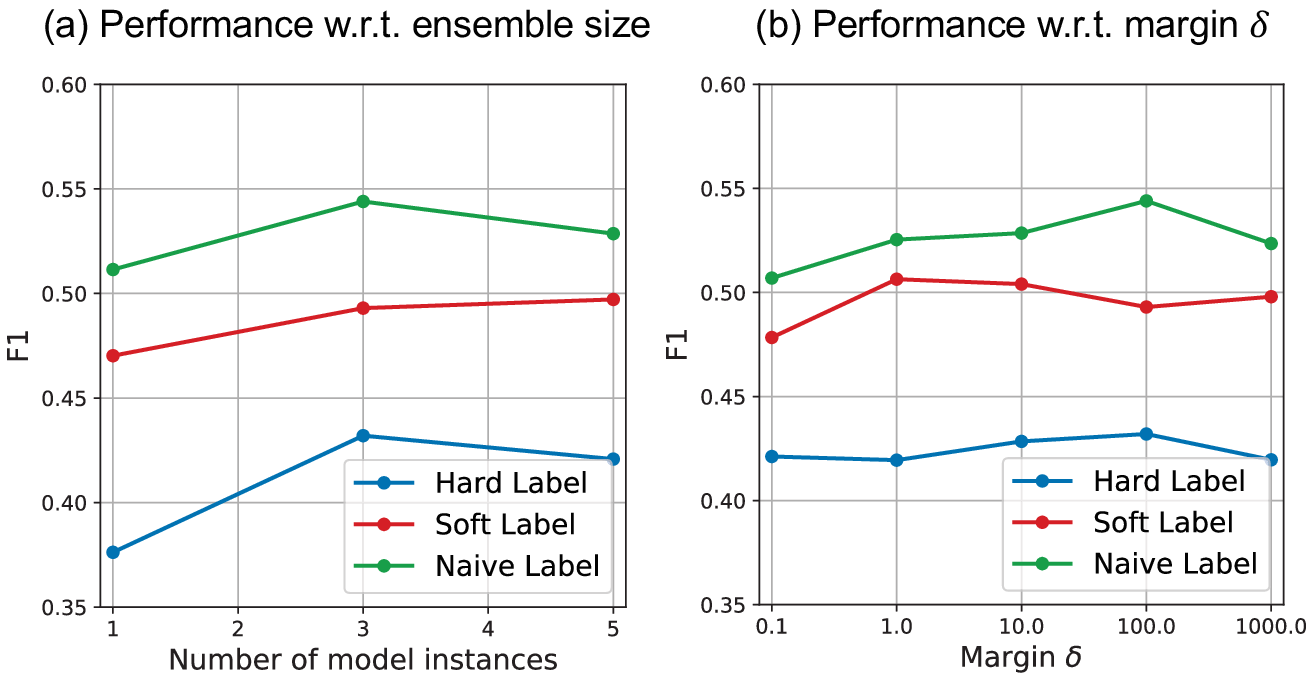}
\caption{The performance of \method-S w.r.t. (a) the ensemble size, and (b) the margin $\delta$ in Eq.~\eqref{eq.hinge}.}\label{fig.parameter}
\vspace{-0.5cm}
\end{figure}

There are two major hyperparameters in \method, the number of model instances in the hybrid and the margin threshold $\delta$ in the hinge loss in Eq.~\eqref{eq.hinge}. In this section, we use \method-S to evaluate the influence of these parameters (\method-M is similar thus is ignored for brevity). Fig. \ref{fig.parameter} presents the change of F1 scores w.r.t. the two parameters on the three labeled sets. In Fig. \ref{fig.parameter}(a), the number represents the the number of either SemiDOC or LightGBM, {\em e.g.}, 3 indicates 3 SemiDOC + 3 LightGBM. We can see that using either 3 or 5 is better than 1, indicating \method-S also benefits from ensembling. Also, the performance marginally improves or degrades after the number is larger than 3, validating our choice in Sec. \ref{sec.setup}. From Fig. \ref{fig.parameter}(b), we can see \method-S is not very sensitive to $\delta$. A small margin ({\em e.g.}, 0.1) is insufficient for distinguishing normal and abnormal entities, and a too large margin ({\em e.g.}, 1000) may lead to overfitting. Thus a proper choice is $\delta=100$, which is our setup for \method.

% \avkan{}{Use relative units in Fig.3 to avoid revealing actual F1 scores.}

\subsubsection{Visualization of Embeddings}

\begin{figure}[!t]
\centering
\includegraphics[width=0.97\columnwidth]{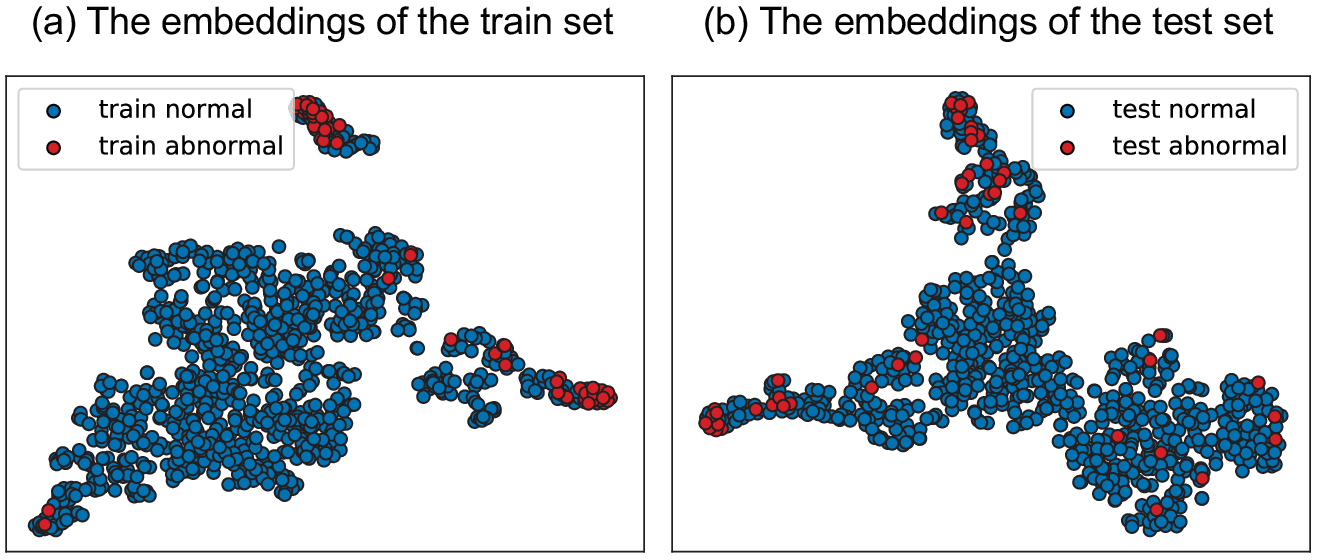}
\caption{The tSNE visualization of the embeddings of SemiDOC using Hard Labels on (a) the train set, (b) the test set.}\label{fig.emb}
\end{figure}

As discussed in Sec. \ref{sec.semioc}, SemiDOC uses a small amount of labeled anomalies to tighten the boundary of the hypersphere of normal entities. To understand how it works, we uniformly sampled a batch of training data and took all test data for visualization (the full training set is too large to visualize). The embeddings of SemiDOC were visualized using tSNE \cite{van2008visualizing} in a 2D space. Fig. \ref{fig.emb} presents the distribution of the training and test embeddings. From Fig. \ref{fig.emb}(a), SemiDOC effectively pushed anomalies away from the boundary of normal entities. As designed by the negative sampling in Eq.~\eqref{eq.semioc}, the anomalies do not need to form a cluster but are used to shape the boundary of normal class. From Fig. \ref{fig.emb}(b), SemiDOC generalizes well on the test set. Although the anomalies overlap with a small amount of normal entities, the majority of the normal class is distinguishable. This explains the design of \method-S, where SemiDOC first filters out most of the normal entities (with a few false negatives), and LightGBM classifies the rest set of normal and abnormal entities.

% \avkan{}{Use more contrasting colors, e.g., blue and orange}

\subsubsection{Ablation Analysis}\label{sec.ablation}

\begin{table}[!t]
\caption{The ablation analysis in terms of F1 score (relative change w.r.t. \method-S). H-Dist. is Hamming distance. E-Dist. is Euclidean distance.}\label{tab.ablation}
\vspace{-0.3cm}
\begin{center}
\small
\begin{tabular}{l|lll}\hline
{\bf Model} & {\bf Hard} & {\bf Soft} & {\bf Naive} \\ \hline
\method-S & {\bf 0.432} & {\bf 0.493} & {\bf 0.544}\\
(a) - Rule features & 0.366 (-15.3\%) & 0.456 (-7.5\%) & 0.497 (-8.6\%)\\
(b) - Alg. features & 0.384 (-11.1\%) & 0.490 (-0.6\%) & 0.521 (-4.2\%)\\
(c) - Meta features & 0.406 (-6.0\%) & 0.485 (-1.6\%) & 0.525 (-3.5\%)\\
(d) SemiOC$\rightarrow$DSVDD & 0.406 (-6.0\%) & 0.480 (-2.6\%) & 0.515 (-5.3\%)\\
(e) SemiOC$\rightarrow$DSAD & 0.403 (-6.7\%) & 0.490 (-0.6\%) & 0.524 (-3.7\%)\\
(f) H-Dist.$\rightarrow$E-Dist. & 0.404 (-6.5\%) & 0.200 (-59.4\%) & 0.437 (-19.7\%)\\ \hline
\end{tabular}
\end{center}
\end{table}

\iffalse{

\begin{table}[!t]
\caption{The ablation analysis in terms of F1 score (relative change w.r.t. \method-S). H-Dist. is Hamming distance. E-Dist. is Euclidean distance.}\label{tab.ablation}
\vspace{-0.3cm}
\begin{center}
\small
\begin{tabular}{l|lll}\hline
{\bf Model} & {\bf Hard} & {\bf Soft} & {\bf Naive} \\ \hline
\method-S & {\bf 0.432} & {\bf 0.493} & {\bf 0.544}\\
(a) - Rule features & 0.366 (+18.0\%) & 0.456 (+8.1\%) & 0.497 (+9.5\%)\\
(b) - Alg. features & 0.384 (+12.5\%) & 0.490 (+0.6\%) & 0.521 (+4.4\%)\\
(c) - Meta features & 0.406 (+6.4\%) & 0.485 (+1.6\%) & 0.525 (+3.6\%)\\
(d) SemiOC$\rightarrow$DSVDD & 0.406 (+6.4\%) & 0.480 (+2.7\%) & 0.515 (+5.6\%)\\
(e) SemiOC$\rightarrow$DSAD & 0.403 (+7.2\%) & 0.490 (+0.6\%) & 0.524 (+3.8\%)\\
(f) H-Dist.$\rightarrow$E-Dist. & 0.404 (+6.9\%) & 0.200 (+146.5\%) & 0.437 (+24.5\%)\\ \hline
\end{tabular}
\end{center}
\end{table}

}\fi

% \avkan{}{Don't reveal actual metric values in the results table. Missing reference.}

In this section, we evaluate several variants of \method-S to validate the design choices of its \fe\ and \ad\ modules. Table \ref{tab.ablation} summarizes the testing results of our ablation analysis. In (a)-(c), we alternately removed the three featurizers in \fe\ (Sec. \ref{sec.featurizer}). In (d) and (e), we replaced SemiDOC in \ad\ module by DeepSVDD and DeepSAD, respectively. In (f), we replaced the Hamming distance with Euclidean distance in Eq.~\eqref{eq.semioc} and Eq.~\eqref{eq.hinge} to evaluate the choice of distance function. First, in (a)-(c), we observe removing any of the featurizers degrades the performance of \method-S. Removing Rule/Algorithmic featurizers have more impacts than the Meta featurizer because they are more fine-grained and dynamic at the time series level. In (d)(e), we observe SemiDOC is better than DeepSVDD and DeepSAD for anomaly detection, validating the design of the negative sampling based regularization in Eq.~\eqref{eq.hinge}. Finally, (f) suggests Hamming distance, which is commonly used in embedding retrieval \cite{song2018deep}, is more suitable than Euclidean distance in \method-S for entity-level anomaly detection. Therefore, the results in Table \ref{tab.ablation} justify the design choices of our method.

\iffalse{

\begin{table}[h!]
\centering
\begin{tabular}{lcccc}
\textbf{Deployment} & \textbf{Precision} & \textbf{Recall} & \textbf{F1} & \textbf{FPR}\\
\hline
Human Labels & 0.45 & 0.27 & 0.34 & 0.07 \\
Bot Labels & - & - & - & 0.02 \\
\hline
Union & 0.33 & 0.27 & 0.29 & 0.03 \\
\hline
\\
\hline
Random Baseline & 0.07 & 0.06 & 0.07 & 0.06 \\
\hline
\end{tabular}
\caption{Standalone performance metrics of the ML component of \method{} algorithm at an aggregated level and per type of labeled deployment. Random Baseline is a Bernoulli model with probability matching the dataset anomaly proportion (6.4\%).}
\label{tab:results}
\end{table}

}\fi

\subsection{Human Evaluation}

\begin{table}[!h]
\caption{Online evaluation of the proposed model.}\label{tab.ab}
\vspace{-0.3cm}
\begin{center}
\small
\begin{tabular}{l|ll}\hline
{\bf Model} & {\bf \# Deployments} & {\bf Precision} \\ \hline
The Existing Model & 64 & 29.7\%\\
Ours & 64 & 35.9\%\\ \hline
\end{tabular}
\end{center}
\end{table}

\vspace{-0.2cm}

% \avkan{}{Show relative improvements}

% \avkan{}{In Table 4, can you also compare the difference in rollback rate? Even if we don't know recall, we can compare rollback rates and precision.}

To evaluate the impact of the proposed \method\ model in real applications, we run an online A/B experiment, where the control group is a LightGBM based model running in the existing rollback system, and the treatment group is our \method\ based model. For fair comparison, both of the existing model and the proposed model were configured to take 50\% traffic of deployments between Jun. 8, 2023, and Jul. 7, 2023. The user of each deployment can score a rollback using the labeling service described in Sec. \ref{sec.system_architecture}. In this period, 128 scored rollbacks were collected, with 64 for each model. We binarized the scores in the same way as Naive Labels, and calculated the precision of each model in Table \ref{tab.ab}. Recall (and F1) cannot be computed because there is no ground truth for the total number of true positives. Because the scores were provided voluntarily, and users tend to provide feedback for wrong decisions, the precisions in Table \ref{tab.ab} are generally lower than those in Table \ref{tab.results}. However, the higher precision of the proposed model indicates it agrees with the users more often than the existing model in the system, suggesting its effectiveness in the online systems.% \guinetgg{}{I'm a bit concerned this sub-section might be more confusing that anything else. One of the main feedback from AMLC was that evaluation was not satisfactory and you've done a great job at making it rigorous above. It's tricky to know if the benefit of showing that this is deployed in prod with medium to good customer feedback is worthy asking the reader to accept a new configuration of settings, even if it's well argued (single label instead of naive/soft/strong, only precision as metrics and small scale of pts). One may also wonder why is the existing model in the table is not directly described above (is it OFE+LGBM ?).}

\iffalse{

\begin{table}[ht!]
\centering
\begin{tabular}{lcccc}
\textbf{Algorithm} & \textbf{Precision} & \textbf{Recall} & \textbf{F1} & \textbf{FPR}\\
\hline
Audubon & 0.41 & $1^{\star}$ & 0.58 & $1^{\star}$ \\
\method & 0.75 & 0.62 & 0.68 & 0.14 \\
\hline
\end{tabular}
\caption{Comparison of performance metrics of Rules-based Audubon and \method. Given that we used Audubon's rollback as a deployment selection criteria for this set, Audubon recall and FPR are equal to 1.}
\label{tab:results-2}
\end{table}

}\fi

\section{Related Work}\label{sec.related_work}

To the best of our knowledge, this the first work for online detection of anomalies in streaming entities where each entity owns its specific MTS. There are several works claimed entity-level anomaly detection (AD) \cite{su2019robust,huang2022semi}, but their methods aim to detect anomalous time points of the MTS emitted by a few entities such as cloud servers. So they are still point-level methods and cannot be applied to streaming entities due to the non-shareable, entity-specific model in \cite{su2019robust} or the need of poin-level supervision in \cite{huang2022semi}.

Typical AD methods include TS-based methods and general methods that can be applied to non-TS vectorial data such as images, as summarized in surveys \cite{schmidl2022anomaly} and \cite{han2022adbench}, respectively. Among them, MTS-based AD methods are most relevant, most of which were designed to be unsupervised. Traditional MTS methods apply general AD techniques such as LOF and iForest on subsequences of MTS. Recent works use RNN or Transformer to encode MTS, and developed reconstruction based methods \cite{park2018multimodal,su2019robust,xu2021anomaly} forecasting based methods \cite{malhotra2015long,munir2018deepant,hundman2018detecting}, and generative methods \cite{li2023prototype}. Moreover, graph-based AD methods have been proposed on MTS by encoding and tracking the change in the correlations between variates using GNN \cite{zhao2020multivariate,deng2021graph} or CNN \cite{zhang2019deep}. However, these methods cannot readily take the advantage of labels, if available. Several recent TS methods were proposed for semi-supervised AD \cite{jiang2021semi,chen2023semisupervised}, but they assume point-level labels are available for locating anomalous time points or segments. This is not the case for entity-level AD as described in Challenge 4 in Sec. \ref{sec.intro}, where labels only mark anomalous entities. Because none of the aforementioned methods can be used to address the challenges of entity-level AD as discussed in Sec. \ref{sec.intro}, a proper solution such as the proposed \method\ is in demand.% In Sec. \ref{sec.exp}, we compare our method with the SOTA point-level MTS methods.

As suggested by \cite{han2022adbench}, leveraging a small amount of labels for semi-supervised AD has a big potential. Recently, general semi-supervised AD methods have been proposed for images \cite{ruff2019deep,huang2021esad}, texts \cite{zhou2023anoonly}, and tabular data \cite{yoon2022spade}. They cannot be used for solving our problem until being embedded into our \method\ framework, such as DeepSAD \cite{ruff2019deep} in Table \ref{tab.ablation}. This suggests the importance and flexibility of the entire \method\ framework for entity-level AD.

\section{Conclusion}\label{sec.conclusion}

In this paper, we introduced \method, a semi-supervised framework for online entity-level anomaly detection. \method\ uses an online feature extractor to align the MTS of different entities to the same feature space, and a hybrid model \ad\ for detecting anomalous entities. In \ad, SemiDOC was proposed for tightening the boundary of normal entities by negative sampling. It is combined with a supervised detector for robust detection. The comprehensive experiments on large-scale datasets indicate \method\ outperforms the SOTA methods, and the human evaluation further suggest its effectiveness in large online systems.

%\begin{acks}
%To Robert, for the bagels and explaining CMYK and color spaces.
%\end{acks}

\clearpage

\bibliographystyle{ACM-Reference-Format}
\bibliography{ref}

\clearpage

\appendix
\section{Appendix}

\subsection{Time Complexity Analysis}\label{sec.complexity}

As an online anomaly detection system, the efficiency for online inference is important. For a new deployment, the computational load of \method\ for online anomaly detection includes the computation for featurizer instances (Sec. \ref{sec.featurizer}) and the inference using \ad\ model. In this section, we first analyze the time complexity for initializing each of the featurizers in Sec. \ref{sec.featurizer} and their score computation, then analyze the time complexity for \ad\ model inference.

\vspace{0.1cm}

\noindent{\bf Rule-based Featurizer.} At the initialization stage, SbF learns the mean $\mu$ and standard deviation $\sigma$ of all $T$ values in the history $\mat{x}_{i,j}$, so its complexity is $O(1)$. TbF sets a predefined threshold $\tau$ with complexity $O(1)$. CbF sets up a sliding window of size $w_{c}$ with complexity is $O(1)$. At inference stage, SbF and TbF check whether $x_{i,j}^{t}$ ($t>T$) is above their thresholds, so the complexity is $O(1)$. Similarly, CbF checks whether $x_{i,j}^{t}$ is a missing value with complexity $O(1)$. Therefore, the complexity of Rule-based Featurizer is $O(1)$.

\vspace{0.1cm}

\noindent{\bf Algorithm-based Featurizer.} At the initialization stage, SubNN segments the length-$T$ history $\mat{x}_{i,j}$ into subsequences with window size $w_{n}$ and stride 1 to form a set $\mathcal{S}$ of length-$w_{n}$ subsequences, so its complexity is $O(T-w_{n})$. MD only sets up a sliding window of size $w_{m}$, so its complexity is $O(1)$. At inference stage, subNN calculates the distance between a length-$w_{n}$ sliding window ending at time step $t$ and its nearest neighbor in $\mathcal{S}$, so the complexity is $O(Tw_{n})$. MD calculates $m_{\text{obs}}=\text{median}([x_{i,j}^{t-w_{m}}, ..., x_{i,j}^{t-1}])$, and $m_{\text{dif}}=\text{median}([x_{i,j}^{t-w_{m}+1} - x_{i,j}^{t-w_{m}}, ..., x_{i,j}^{t-1} - x_{i,j}^{t-2}])$ with complexity $O(w_{m})$. Therefore, the complexity of Algorithm-based Featurizer is $O(Tw_{n} + w_{m})$.

\vspace{0.1cm}

\noindent{\bf Meta-Data Featurizer.} The deployment-level meta-data features are static and can be retrieved and appended to rule-based and algorithm-based features with $O(1)$ time complexity.

\vspace{0.1cm}

On top of the featurizers, there is a time pooling layer (Sec. \ref{sec.time_pooling_layer}) for addressing the variable duration of different deployments and a feature aggregator (Sec. \ref{sec.aggregator}) for aligning different deployments into the same feature space. For the time pooling layer, both MaxPool and MeanPool in Eq.~\eqref{eq.pooling} were updated incrementally in constant time during online inference, and in parallel for different features, so its complexity is $O(1)$. For feature aggregator, the key computational step in Eq.~\eqref{eq.agg} was implemented in parallel for all metrics, and its time complexity for each metric is $O(n_{i})$ for $n_{i}$ variates in the MTS. Additionally, the imputation step in the feature aggregator takes $O(n_{i})$ time for filling missing values at a time step. Therefore, by parallelizing all featurizer instances, the overall time complexity of the \fe\ module during online inference is $O(Tw_{n} + w_{m} + n_{i})$.

\vspace{0.1cm}

\noindent{\bf \ad\ Model Inference.} The SemiDOC module in \ad\ uses Eq.~\eqref{eq.semiocscore} for inferring anomaly scores, where $\mat{c}$ and $R$ were pre-computed during training time and stored for model inference. So the key computation is the embedding $\phi(\mat{z}_{\text{new}}; \boldsymbol{\theta})$, which takes $O(dd_{\text{max}})$ for an MLP encoder with $d_{\text{max}}$ as the maximal dimension of all layers, where $d$ is the dimension of $\mat{z}_{\text{new}}$. In addition, the LightGBM in \ad\ uses $O(d)$ time for inference with a constant number of estimators \cite{ke2017lightgbm}. Therefore, the time complexity for both ensemble methods \method-M and \method-S are $O(dd_{\text{max}})$ at online inference time.

\vspace{0.1cm}

\noindent{\bf Summary.} Integrating the time complexity of the processes for initializing featurizers, online computation of features, and \ad\ model inference, the time complexity of \method\ for online anomaly detection is $O(Tw_{n} + w_{m} + dd_{\text{max}})$, where $T$ is the length of historical time series, $w_{n}$ is the window size used by SubNN, $w_{m}$ is the window size used by MD, $d$ is the dimension of features input to SemiOC, and $d_{\text{max}}$ is the maximal dimension of all layers in the neural network $\phi(\cdot; \boldsymbol{\theta})$. In practice, $w_{n}$, $w_{m}$, and $d_{\text{max}}$ are set as small constants. According to Sec. \ref{sec.aggregator}, $d=134$ is also a small constant. Therefore, the time complexity is approximately $O(T)$. In our experiments, we used 2-day historical time series in minute-wise, so $T=2880$. We set $w_{n}=100$, $w_{m}=100$, and $d_{\text{max}}=128$. According to Sec. \ref{sec.aggregator}, $d=134$. With this setup, the model is efficient and can perform online anomaly detection.

\subsection{Implementation Details}\label{sec.implementation}

For the baseline methods, we employed their official code when available. In the group of MTS-based methods, OmniAnomaly used a 2-layer GRU, 3-layer encoder, and 3-layer decoder, with PReLU as the internal activation and Sigmoid as the output activation. Its embedding dimension was set as 8, and other hidden layers had dimension of 32. Its hyperparameter $\beta=0.01$ for KL divergence. It was trained with Adam optimizer \cite{kingma2014adam} with learning rate $2e^{-3}$ and weight decay $1e^{-5}$. AnomalyTransformer used a 3-layer transformer with 8 heads as the encoder. The embedding dimension was set as 512, and the activation function was GELU. Its hyperparameter $\lambda=3$. I was trained with Adam optimizer with learning rate $1e^{-4}$. TranAD used a transformer encoder-decoder architecture with embedding dimension of 16, appended with a fully connected output layer. It was trained with Adam optimizer with learning rate $1e^{-3}$ and weight decay $1e^{-5}$.

In the group of general methods, both DeepSVDD and DeepSAD used a 2-layer neural network encoder with embedding dimension 128. LeakyRELU (slope 0.1) was used as the activation function in the input layers, and Sigmoid was used as the output activation. LayerNorm \cite{ba2016layer} was added in each layer. The hyperparameter of DeepSAD was set as $\eta=1$. Both DeepSVDD and DeepSAD were optimized by Adam with learning rate $1e^{-3}$, weight decay $1e{-4}$. RandomForest used 100 estimators with maximum depth 2. LightGBM used 100 estimators with maximum depth as 5, and a learning rate of 0.1.

For the proposed \method\ method, the neural network $\phi(\cdot; \boldsymbol{\theta})$ in SemiDOC in Eq.~\eqref{eq.semioc} was implemented with a two layer encoder with embedding dimension 128. LeakyRELU (slope 0.1) was used as the activation function in the input layers, and Sigmoid was used as the output activation. LayerNorm was added in each layer. The architecture was the same as DeepSVDD and DeepSAD for fair comparison. The SemiDOC model was optimized by Adam with learning rate $1e^{-3}$, weight decay $1e^{-4}$, batch size 256, and a maximum of 500 epochs. Early stopping was employed using the validation set.

\end{document}